\documentclass[letterpaper, 10 pt, conference]{ieeeconf}  

\IEEEoverridecommandlockouts                              

\usepackage{bbm}
\usepackage{amsmath}
\DeclareMathOperator*{\argmax}{arg\,max}
\DeclareMathOperator*{\argmin}{arg\,min}
\usepackage[ruled, lined, noend, linesnumbered]{algorithm2e}
\usepackage{booktabs}
\usepackage{graphicx}
\usepackage{subfigure}
\usepackage{makecell}
\usepackage{multirow}
\usepackage{cite}
\usepackage{xcolor}
\let\labelindent\relax
\usepackage{enumitem}

\def\x{\mathcal X}

\title{\LARGE \bf
Robot Motion Planning as Video Prediction: A Spatio-Temporal Neural Network-based Motion Planner}

\author{Xiao Zang$^1$\quad Miao Yin$^1$\quad Lingyi Huang$^1$\quad Jingjin Yu$^2$\quad Saman Zonouz$^1$\quad Bo Yuan$^1$ 
\thanks{$^{1}$Department of Electrical and Computer Engineering, Rutgers University. $^{2}$Department of Computer Science, Rutgers University. This work is supported by the National Science Foundation (NSF) under Grant CCF-1937403 and CNS-1703782.}%
}

\begin{document}

\maketitle
\thispagestyle{empty}
\pagestyle{empty}

\begin{abstract}
Neural network (NN)-based methods have emerged as an attractive approach for robot motion planning due to strong learning capabilities of NN models and their inherently high parallelism. Despite the current development in this direction, the efficient capture and processing of important sequential and spatial information, in a direct and simultaneous way, is still relatively under-explored. 
To overcome the challenge and unlock the potentials of neural networks for motion planning tasks, in this paper, we propose STP-Net, an end-to-end learning framework that can fully extract and leverage important spatio-temporal information to form an efficient neural motion planner. By interpreting the movement of the robot as a video clip, robot motion planning is transformed to a video prediction task that can be performed by STP-Net in both spatially and temporally efficient ways.
Empirical evaluations across different seen and unseen environments show that, with nearly 100\% accuracy (aka, success rate), STP-Net demonstrates very promising performance with respect to both planning speed and path cost. Compared with existing NN-based motion planners, STP-Net achieves at least $5\times$, $2.6\times$ and $1.8\times$ faster speed with lower path cost on 2D Random Forest, 2D Maze and 3D Random Forest environments, respectively. Furthermore, STP-Net can quickly and simultaneously compute multiple near-optimal paths in multi-robot motion planning tasks.
\end{abstract}

\section{Introduction}
Sampling-based motion planners (SMP) \cite{kuffner2000rrt}\cite{kavraki1996probabilistic} provide an effective solution for finding collision-free paths with probabilistic completeness guarantees. However, existing asymptotically optimal SMPs, e.g., RRT*~\cite{karaman2011sampling} and its many variants, can take significant computational time as their solutions converge to optimal ones. A main reason for this drawback 
is due to the sampling and collision checking procedures being sequential, which leads to high exploration cost in the planning phase. 
To address the shortcoming, a potentially promising solution is to use a biased sampling scheme that can always guide the robot to move to the next waypoint lying on the shortest collision-free path. Evidently, such a scheme, properly designed, e.g., through introducing parallelism, can reduce the planning time for finding optimal or near-optimal paths significantly.


Motivated by this philosophy, recently, a number of learning-based SMPs have been proposed to learn proper biased sampling schemes from data \cite{qureshi2019motion,ichter2019robot,havens2019learning,ichter2018learning,chen2019learning,toma2021waypoint,inoue2019robot}, and most of them adopt neural networks (NNs) to guide the sampling process. Thanks to the inherently high parallelism of NNs and their powerful learning capabilities, these emerging data-driven SMPs have already demonstrated promising planning performance in both time cost and path cost.

Despite their success, from both temporal and spatial perspectives, state-of-the-art NN-based motion planners face two limitations. \underline{First}, it is unclear that important sequential structures in motion planning tasks have been fully extracted or learned in current NN-based planners. In other words, motion planning problems have an inherently sequential nature in that they look for a \emph{path} that marches from the start to the goal. However, many existing NN-based motion planners (e.g., MPNet~\cite{qureshi2020motion} and L-SBMP~\cite{ichter2019robot}) adopt either multi-layer perceptron (MLP) or convolutional neural network (CNN) as the backbone network architecture, which by design are not optimal choices for processing sequential information. \underline{Second}, rich spatial correlation in the environment is not properly modeled or leveraged. To be specific, in order to integrate the information of the environment and the robot state into the learning procedure, many existing NN-based works \cite{toma2021waypoint}\cite{watt2020pathnet}\cite{tranpredicting} encode the high dimensional workspace into low dimensional vector representation via contractive auto-encoders (CAE)~\cite{rifai2011contractive}. Such straightforward embedding strategy typically cannot fully extract the important spatial correlation between environment and robot state, thereby potentially severely hampering the planning performance.

\begin{figure}[t]
\vspace{1mm}
    \centering
    \includegraphics[width=\linewidth]{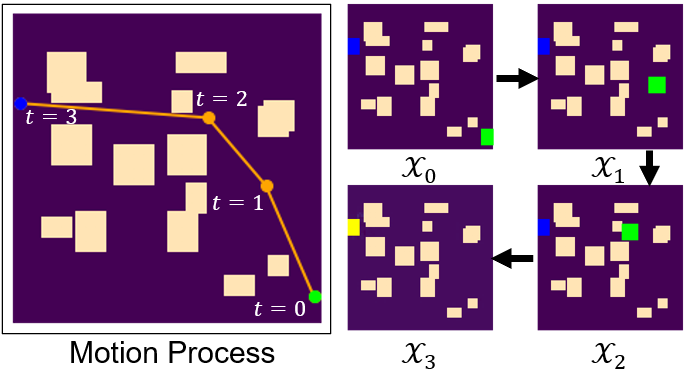}
    \vspace{-7.5mm}
    \caption{An example of a 2D planning task from a video prediction perspective. The left figure shows the generated path, where the white blocks indicate the obstacle space and the green/blue dots represent the start/goal states. The right figures are the corresponding video clip, where each frame corresponds to a discrete observation of the motion process. Current and goal states are represented as $5 \times 5$ patches to augment the state information.}
    \label{fig:videoclip}
    \vspace{-2.5mm}
\end{figure}

Toward overcoming these limitations, we propose an end-to-end learning framework that can fully extract and leverage important spatio-temporal information to form an efficient NN-based motion planner. Our key idea is to interpret the full motion sequence of a robot as a video clip composed of a series of sequential discrete snapshots of the statuses of the planning task, across different timestamps (see Figure~\ref{fig:videoclip}). From this perspective, robot motion planning can be then transformed to a video prediction task. We propose \textbf{STP-Net}, a spatio-temporal neural network model that can efficiently capture and process the spatial correlation and temporal dependence \textbf{simultaneously and directly from the data sequence}, thereby bringing high planning performance. The contributions of this paper are summarized as follows:
\begin{itemize}[leftmargin=3.5mm]
    \item We propose to interpret robot motion planning as a video prediction task. By delicately mapping both the environment and start/goal states into the different channels of colored video clips, important spatial information can be maximally preserved and properly incorporated into the neural network model.
    
    \item We propose STP-Net, an end-to-end neural network architecture that can directly learn from video data without suffering encoding loss. At the same time, the spatio-temporal processing capability of STP-Net enables the simultaneous identification and extraction of both spatial environment information and temporal movement correlations, thereby yielding high-quality predictions. 
    
    \item We perform empirical evaluations of STP-Net across different seen and unseen environments. With close to 100\% accuracy, STP-Net demonstrates promising performance with respect to the planning speed and path cost, across many different environments. 
    
    \item We further extend our STP-Net to solving multi-robot path planing problems. Empirical results confirm the decent scalability of STP-Net in computing multiple near-optimal paths simultaneously and quickly, without degradation in planning accuracy.

\end{itemize}


\section{Related Work}
RRT~\cite{lavalle1998rapidly} builds a tree structure that rapidly explores the configuration space by repeatedly connecting uniformly generated samples, until a feasible path can be extracted from the tree. 
RRT*~\cite{karaman2011sampling} extends RRT by incrementally rewiring the RRT tree branches, thereby reducing the length of the computed path. Due to the curse of dimensionality, RRT* suffers from slow convergence to find optimal solutions, e.g., for a high-DOF robot or a multi-robot system. 
Aiming to improve both the convergence rate and path quality of RRT*, Informed-RRT*~\cite{gammell2014informed} is proposed to define an ellipsoidal search region of new samples, based on an initial solution from RRT*. Though Informed-RRT* demonstrates enhanced convergence to an optimal solution, it is still limited by the time cost of finding an initial solution. To address this issue, Batch Informed Trees (BIT*)~\cite{doi:10.1177/0278364919890396} is further designed to unify graph search technique and SMP, which outperforms both RRT* and Informed-RRT*. 

Recently, several NN-powered motion planners have been developed and reported to predict the future states or trajectories directly. Among these, DeepSMP~\cite{qureshi2018deeply} uses a contractive auto-encoder (CAE) to pre-encode the obstacle point clouds into a vectorized representation, which can be then concatenated with the current and goal states of robot. Such combined vector is then fed to a multi-layer perceptron (MLP) for the prediction of the next state. MPNet~\cite{qureshi2020motion} shares the similar design strategy that DeepSMP adopts, and it is further equipped with re-planning and rewiring mechanisms to improve performance. With the same input as MPNet uses, PathNet~\cite{watt2020pathnet} predicts the future trajectory instead of the state. L-SBMP~\cite{ichter2019robot} also utilizes CAE to embed the high-dimensional environmental information in a low-dimensional latent space. Different from DeepSMP and MPNet that perform combination of real states and environments in the latent space, L-SBMP chooses to combine them first and then encode the combination. On the other hand, \cite{khan2020graph} integrates the graph embedding into the variational auto-encoder (VAE)~\cite{kingma2013auto} to identify the critical node in an iterative manner. To better capture sequential information, OracleNet~\cite{bency2019neural} proposes to leverage long short-term memory (LSTM) to learn the sequences of states along the paths. However, its performance is limited due to the insufficient integration of environmental information.

Instead of predicting the specific states, some other works predict the sampling distribution where the near-optimal paths exist. For example, \cite{ichter2018learning} and \cite{kumar2019lego} use conditional VAE to predict the sample distribution. 
Besides, \cite{shah2020learning} identifies regions on the map images via the convolutional neural network to bias the sampling. In addition, \cite{yu2021reducing} adopts message passing algorithm to predict the exploration priority in the random geometric graph, thereby facilitating the generation of the feasible paths. 



\section{Preliminaries }
\label{sec:prob}

\textbf{Notation.} Let $\mathcal{C}$ denote the configuration space (C-space) of a robot, where the obstacle and free regions are denoted as $\mathcal{C}_{obs}$ and $\mathcal{C}_{free} = \mathcal{C} \setminus \mathcal{C}_{obs}$, respectively. Given the start state $x_{start} \in \mathcal{C}_{free}$ and the goal state $x_{goal} \in \mathcal{C}_{free}$, the objective of motion planning is to find a collision-free path between $x_{start}$ and $x_{goal}$ in $\mathcal{C}_{free}$. Let $x$ denote a discrete sequence of waypoints lying on a path, where $x(0)=x_{start}$ and $x(T) = x_{goal}$. The waypoint sequence $x$ is valid if $x(t) \in \mathcal{C}_{free}, \forall t \in (0,T)$, and the path connecting all continuous states in $x$ lies entirely in the free space $C_{free}$.


\textbf{Environment Representation.} For a 2D or 3D environment, given the spatial resolution, its map and the start/goal configuration can be represented using the 0/1-valued pixels or voxels. For instance, a square 2D environment can be described by an $n \times n \times 3$  image. Here, the information of free space and obstacles are represented as zero-valued and one-valued pixels in the first channel of the image, respectively. The second (resp., third) channel marks the pixels that correspond to the start (resp., goal) position as ones while all the other pixels channels take zero values. Following a similar representation scheme, a 3D environment can be encoded by an $n \times n \times n \times 3$ array of voxels.

\begin{figure}[ht]
    \centering
    \includegraphics[width=\linewidth]{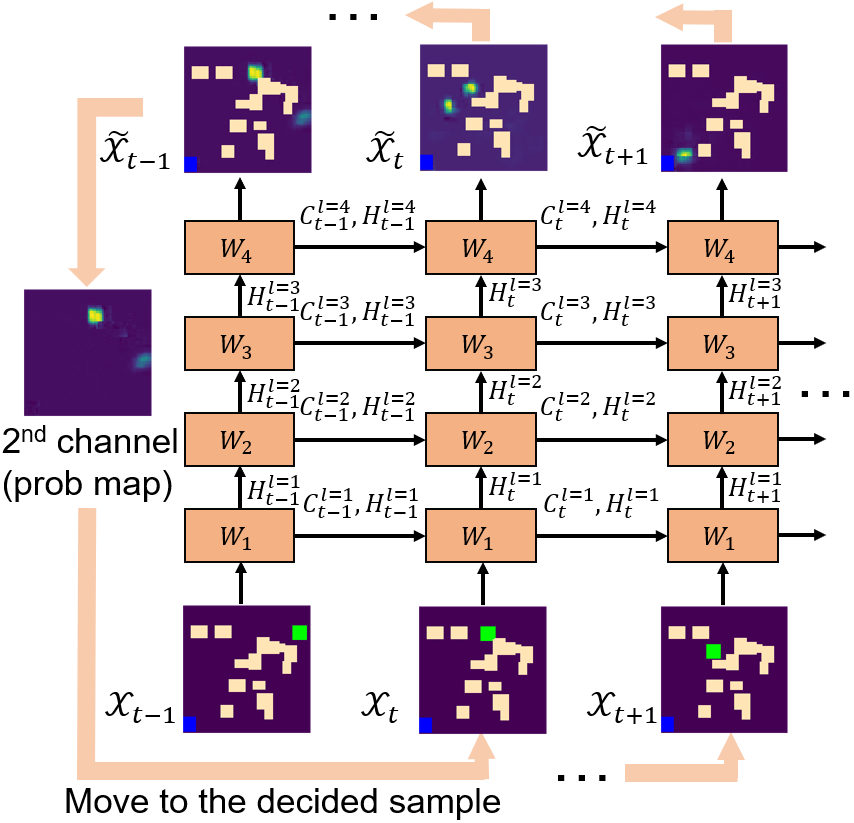}
    \vspace{-6.5mm}
    \caption{The processing procedure of STP-Net when being unfolded along the time dimension. The weights $W_{1-4}$ are shared across different timestamps. The network takes the current video clip as input, and predicts the next video frame whose second channel corresponds to the probability distribution of the next waypoint.}
    \label{fig:convlstm}
    \vspace{-3mm}
\end{figure}

\section{Methods}
\subsection{Robot Motion Planning: A Video Prediction Perspective}

\textbf{Interpreting Robot Motion as Video.} Based on the above described environment representation scheme, the discrete process of robot motion planning can be viewed as composing a video clip. 
Suppose we are monitoring the discrete motion of a robot in a 2D/3D environment, where the robot keeps moving from the start state to a fixed goal state. The full set of discrete statuses of the planning task can be then viewed as a sequence of video frames $\mathcal{X}_0,\mathcal{X}_1,...,\mathcal{X}_{T}$, where the first frame $\mathcal{X}_0$ corresponds to the initial status of the task and the last frame $\mathcal{X}_T$ describes the final status of the task (see Figure \ref{fig:videoclip}). Noticing that a planned path is essentially the sequential connection of multiple waypoints $x(t)$, we make the design choice that each video frame $\mathcal{X}_t$ represents the spatial information of $x(t)$ together with the corresponding environment and goal state $x(T)$ at timestamp $t$. 

\textbf{Pixel Patching.} Recall that in principle, the robot state and the goal state can be represented using only one pixel in the second and third channels, respectively. 
However, in practice, NN models may not be able to capture and learn such one-pixel information efficiently, and thereby potentially affecting planning performance. To address this issue, we propose to augment the representation of the state information using more pixels. To be specific, we mark all the free-space pixels within the patches of size $p$ centered on the robot state pixel and goal state pixel as ones, respectively. Figure~\ref{fig:videoclip} provides an example of the patch augmentation. In general, a large $p$ reduces the state resolution, while a small $p$ increases the difficulty of the model to capture the state representation. Therefore, the value of patch size $p$ is properly determined by balancing such trade-offs in our experiments.



\textbf{Performing Motion Planning as Video Prediction.} With such video-based interpretation, the path finding in 2D/3D workspace is essentially equivalent to a video prediction task. In general, given the current and previous observations $\mathcal{X}_0,\mathcal{X}_1,...,\mathcal{X}_{t}$ at timestamp $t$, our goal is to predict the new frame $\mathcal{X}_{t+1}$, whose second channel indicates a sample map of the next waypoint $x(t+1)$ with probabilities in $[0, 1]$ (see Figure~\ref{fig:convlstm}). This process is repeated until the next waypoint reaches the goal state $x(T)$. Notice that here we aim to find the path with minimal Euclidean distance, as measured by the cost function
\begin{align}\label{eq:cost}
\mathcal J(\x_0, \ldots, \x_T) = \sum_{t = 1}^{T} \| x(t) - x(t -1) \|_2.
\end{align}

\subsection{Building End-to-End Spatio-Temporal Network}
\label{sec:nn}

Inspired by the current success of neural networks in video processing applications, we propose to build an end-to-end spatio-temporal neural network to perform this video prediction-based motion planning task. Here, distinct from existing CAE-equipped neural motion planners, our proposed network model directly uses the raw spatial information as the input without any CAE-based encoding. 
Specifically, for our target motion planning task, if the neural network is properly designed that they are able to directly process the non-encoded environmental and state information, the originally inevitable information loss incurred by CAE encoding can be completely avoided, thereby bringing better planning performance. In such scenarios, the neural network can perform end-to-end learning to predict the most probable length-$K$ sequences of video frames in the future given the previous $J$ observations as follows: 
\begin{equation}
\label{eq:train}
    \tilde{\mathcal{X}}_{t+1:t+K} = \argmax_{\tilde{\mathcal{X}}_{t+1:t+K}}p(\mathcal{X}_{t+1:t+K}|\hat{\mathcal{X}}_{t-J+1:t}).
\end{equation}

To achieve this goal, we build a spatio-temporal LSTM as the backbone network architecture. As illustrated in Figure~\ref{fig:convlstm}, the entire neural network consists of four ConvLSTM layers~\cite{xingjian2015convolutional}, which can directly process our prepared video frame as the input. The key equations are as follows:
\begin{equation*}
\begin{split}
g_t &= tanh(\mathcal{W}_{xg}*\mathcal{X}_t + \mathcal{W}_{hg} * \mathcal{H}_{t-1} + b_g), \\
i_t &= \sigma(\mathcal{W}_{xi}*\mathcal{X}_t + \mathcal{W}_{hi}*\mathcal{H}_{t-1} + \mathcal{W}_{ci} \odot \mathcal{C}_{t-1} + b_i), \\
f_t &= \sigma(\mathcal{W}_{xf}*\mathcal{X}_t + \mathcal{W}_{hf}*\mathcal{H}_{t-1}+\mathcal{W}_{cf}\odot \mathcal{C}_{t-1} + b_f), \\
\mathcal{C}_t &= f_t \odot \mathcal{C}_{t-1} + i_t \odot g_t, \\
o_t &= \sigma(\mathcal{W}_{xo} * \mathcal{X}_t + \mathcal{W}_{ho} * \mathcal{H}_{t-1} + \mathcal{W}_{co}\odot \mathcal{C}_t + b_o), \\
\mathcal{H}_t &= o_t \odot tanh(\mathcal{C}_t),
\end{split}
\end{equation*}
where $\sigma$, $*$ and $\odot$ represent the sigmoid activation function, the convolution operator and element-wise product, respectively. Here, for 2D tasks, all the inputs $\mathcal{X}_t$, cell outputs $\mathcal{C}_t$, hidden states $\mathcal{H}_t$, and gates $i_t, f_t, g_t, o_t$ are 3D tensors in $\mathbf{R}^{K \times n \times n}$, where $K$ is the number of input dimensions or the number of feature maps (for hidden representations). For 3D tasks, these are 4D tensors in $\mathbf{R}^{K \times n \times n \times n}$ instead. Tensors $\mathcal{W}$ and $b$ are the training parameters, which are represented by $\mathcal{W}_{1-4}$ in Figure~\ref{fig:convlstm} for simplicity of notations.

After the feature extraction at each timestamp , the output layer is connected with an additional convolutional layer to convert the last hidden state into the video frame for prediction results. In other words, the network performs the learning on the input video frame and output the prediction in the same frame format for the next-step planning, thereby fully preserving and leveraging the spatial correlation information at each timestamp in a consistent and end-to-end way. 
At the same time, because our full neural network exhibits LSTM characteristics by design, compared with existing CNN or MLP-based motion planners, our approach can more properly capture and leverage the sequential dependence and correlation among different timestamps. 
All in all, with the carefully designed end-to-end convolutional LSTM architecture, the rich and important spatio-temporal information exhibited in the entire motion planning procedure can be fully identified, extracted and processed, thereby significantly improving the planning performance.

\begin{figure*}[t]
  \centering
  \includegraphics[width=\linewidth]{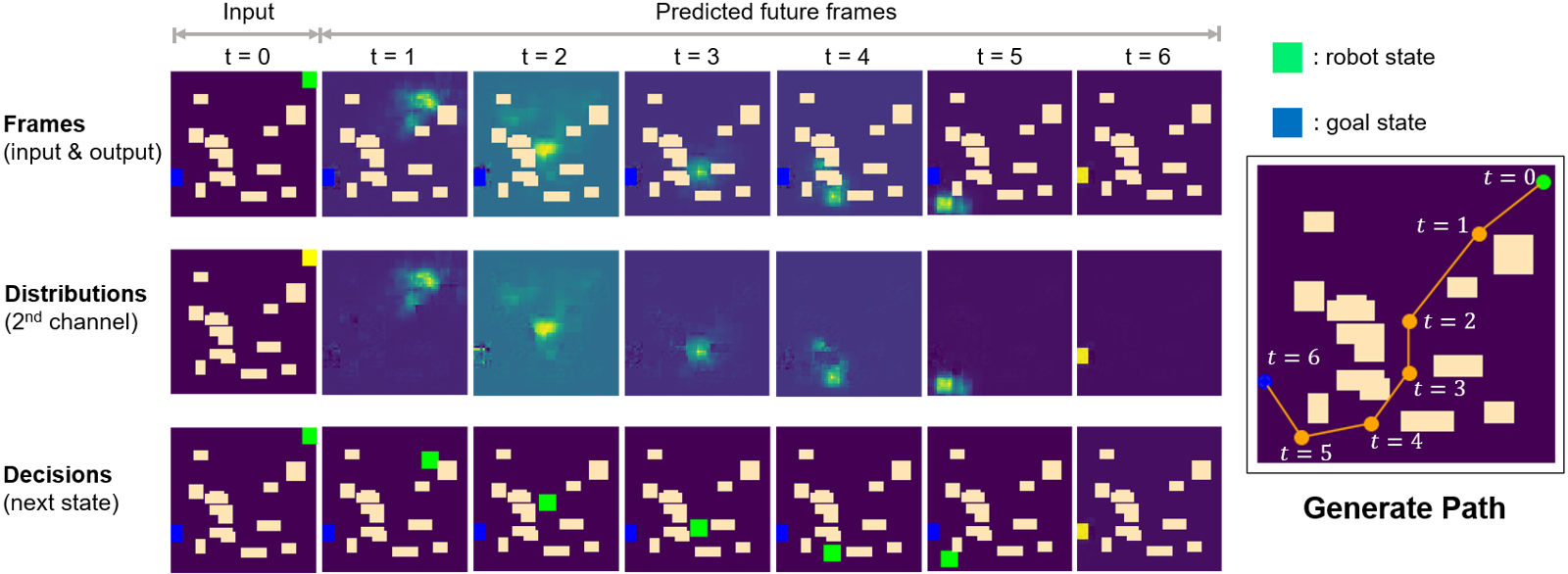}
  \vspace{-7mm}
  \caption{Examples of ST-Net based motion planning in the 2D Random Forest workspace. The figures in the first row ("Frames") are the predicted video frames, whose white, green and blue blocks represent the obstacle space, the regions of the current states and the goal regions, respectively. In the second row ("Distributions"), the sampling distribution of the next state is shown by visualizing the second channel of the predicted frames. The yellower pixel indicates that the corresponding entry has a higher probability of being the next state. The third row ("Decisions") shows the selected next state determined by the predicted sampling distribution and collision detection. Finally, the generated path is shown at the right side.}
  \label{fg:process}
  \vspace{-2mm} 
\end{figure*}

\begin{algorithm}[t]
\begin{small}
  \caption{STP-Net (Online Single-Path Planning)}
  \label{alg:st}
  \DontPrintSemicolon
    \KwIn{The trained NN model $f$, first frame $\mathcal{X}_0$, $x_{start}$, $x_{goal}$, patch size $p$, max iteration $max\_itr$}
    \KwOut{The near-optimal path $\mathcal{P}$}
    The current state $x_{current} \leftarrow x_{start}$ \;
    The observed motion process $\mathcal{V} \leftarrow [\mathcal{X}_0]$ \;
    \For {$itr \leftarrow 0$ \textbf{to} $max\_itr$} {
        $\mathcal{X}_{next} \leftarrow f(\mathcal{V}) $ \;
        $\mathcal{S} \leftarrow argsort(\mathcal{X}_{next}[1])[::-1]$ \;
        \For{$x_{next} \in \mathcal{S}$} {
            \If{$x_{next} \in \mathcal{P}$ or $ColDet(x_{next}, x_{current})$} {
                \textbf{continue} \;
            }
            $x_{current} \leftarrow x_{next}$ \;
            $\mathcal{P}.append(x_{current})$ \;
            \If {$x_{current} \in region(x_{goal})$} {
                $\mathcal{P}.append(x_{goal})$ \;
                \textbf{return} $\mathcal{P}$
            }
            $\mathcal{X}_{new} = GenFrame(\mathcal{X}_0, x_{current}, p)$\;
            $\mathcal{V}.append(\mathcal{X}_{new})$\;
            
            \textbf{break} \;
        }
    }
    \textbf{return} $\O$ \;
\end{small}
\end{algorithm}

\begin{algorithm}[t]
\begin{small}
  \caption{STP-Net (Online Multi-Path Planning)}
  \label{alg:mt}
  \DontPrintSemicolon
    \KwIn{The trained NN model $f$, first frame $\mathcal{X}_0$, the list of start states $x_{start}$, the list of goal states $x_{goal}$, patch size $p$, max iteration $max\_itr$, the number of robots $num\_robot$}
    \KwOut{The list of near-optimal paths $\mathcal{P}$}
    The list of current states $x_{current} \leftarrow x_{start}$ \;
    The observed motion process $\mathcal{V} \leftarrow [\mathcal{X}_0]$ \;
    The solution indicators $Found \leftarrow [False] * num\_robots$ \;
    \For {$itr \leftarrow 0$ \textbf{to} $max\_itr$} {
        $\mathcal{X}_{next} \leftarrow f(\mathcal{V}) $ \;
        \For {$id \leftarrow 0$ \textbf{to} $num\_robot$}{
            \If{$Found[id]$}
            {
                \textbf{continue} 
            }
            $\mathcal{S} \leftarrow argsort(\mathcal{X}_{next}[2*id + 1])[::-1]$ \;
            \For{$x_{next} \in \mathcal{S}$} 
            {
                \If{$x_{next} \in \mathcal{P}[id]$} {
                    \textbf{continue} 
                }
                \If{$ColDet(x_{next}, x_{current}[id])$}
                {
                    \textbf{continue} 
                }
                $Occupied \leftarrow False$ \;
                \For{$j \leftarrow 0$ \textbf{to} $num\_robot$}
                {
                    \If{$x_{next} == x_{current}[id]$}
                    {
                        $Occupied \leftarrow True$
                    }
                }
                \If{$Occupied$}
                {
                    \textbf{continue} 
                }
                $x_{current}[id] \leftarrow x_{next}$ \;
                $\mathcal{P}[id].append(x_{current}[id])$ \;
                \If {$x_{current}[id] \in region(x_{goal}[id])$} {
                    $\mathcal{P}[id].append(x_{goal}[id])$ \;
                    $Found[id] \leftarrow True$
                }
                \textbf{break} 
                
            }
        }
        $\mathcal{X}_{new} = GenFrame(\mathcal{X}_0, x_{current}, p)$\;
        $\mathcal{V}.append(\mathcal{X}_{new})$\;
        $All\_found = True$\;
        \For{$id \leftarrow 0$ \textbf{to} $num\_robot$}
        {
            \If{\textbf{not} $Found[id]$}
            {
                $All\_found\leftarrow False$
            }
        }
        \If{$All\_found$}
        {
            \textbf{return} $\mathcal{P}$
        }
    }
    \textbf{return} $\O$ \;
\end{small}
\end{algorithm}

\subsection{Network Training and Path Construction}

\textbf{Neural Network Training (Offline Phase).} To train the neural network on obtaining path planning capability, we prepare the training dataset via using the discrete motion process created by an ``expert'' planner across different tasks. Specifically, we generate the near-optimal paths using RRT* and extract their waypoints to form the video clips. During training, STP-Net takes the observations over the first $J$ timestamps as the input sequence $\hat{\mathcal{X}}_{in} = \{\hat{\mathcal{X}}_0, \hat{\mathcal{X}}_1,...,\hat{\mathcal{X}}_J\}$, and predicts the remaining frames $\tilde{\mathcal{X}}_{out}=\{\tilde{\mathcal{X}}_{J+1}, \tilde{\mathcal{X}}_{J+2}, ..., \tilde{\mathcal{X}}_{T}\}$. In general, if we denote the function of neural network model $\theta$ as $f_\theta$, the training goal is to find the optimal parameters $\theta^{*}$ that minimizes the average cross-entropy loss between the predicted sequences and true targets across all the training data points $(\{\hat{\mathcal{X}}_{in}, \hat{\mathcal{X}}_{out}\})_n$ as:
\begin{equation}
\theta^{*} = \argmin_{\theta}CrossEntropy(f_{\theta}( \hat{\mathcal{X}}_{in}), \hat{\mathcal{X}}_{out}).
\end{equation}
Here, the standard stochastic gradient descent (SGD)-based training strategy is used to train the desired STP-Net model.



\textbf{Single-Path Construction (Online Phase).}
After the network training phase, during the online path construction we use the well-trained STP-Net to generate the sample distribution for the waypoint. As described in Algorithm~\ref{alg:st}, given a well trained STP-Net model $f$, the initial frame $\mathcal{X}_0$ and the start/goal states, the robot state is first initialized as $x_{start}$ (Line 1), and there is only one frame $\mathcal{X}_0$ in the current observation (Line 2). Then STP-Net performs the prediction in an iterative manner (Line 3). At each iteration, the NN model $f$ takes the newest observation $\mathcal{V}$ and predicts the next frame $\mathcal{X}_{next}$, which shares the same size as the input frame (Line 4). Recall that here the second channel of $\mathcal{X}$ records the current state of the robot, while the other two channels are the static environment and the fixed goal state, respectively. Also, because all the pixels are between $0$ and $1$ which indicates their probability of being the next waypoint, we can then sort the entries of the predicted frame $\mathcal{X}_{next}$ by their pixel values in the second channel (Line 5). In this way, the foremost element in $\mathcal{S}$ has the greatest probability to be the next waypoint. Notice that the elements in $\mathcal{S}$ are sequentially visited and checked if they are not used or collision-free (Line 6-8). Specifically, the function $ColDet$ checks whether the line connecting the current state $x_{current}$ and the next candidate state $x_{next}$ has collision with the obstacles or not. If $x_{next}$ is valid (not used and collision-free), the robot moves to the new state and appends it into the path (Line 9). The path is set to be found once $x_{current}$ reaches the region around the goal state (Line 11-13). Otherwise, we formulate the current task status as the new frame $\mathcal{X}_{new}$ and append it into the observation $\mathcal{V}$ (Line 14-15). The function $GenFrame$ produces the new frame by marking the patch corresponding to $x_{current}$ in the second channel (start state channel), while the other two channels (environment and goal state channel) are exactly the same as the initial frame $\mathcal{X}_0$. If the path still can not be found after $max\_itr$ iterations, an empty solution is returned (Line 17).

\textbf{Extension to Multi-Path Construction.} The proposed STP-Net can also be readily extended to serve as a multi-robot path planner, which is described in Algorithm~\ref{alg:mt}. To that end, for each additional robot under the same environment, the raw input frame $\mathcal{X}_0$ is stacked with two extra channels, which represent the start/goal states of the new robot. For instance, for 2D and 3D planning tasks of five robots, the sizes of input frames are $n \times n \times 11$ and $n \times n \times n\times 11$, respectively. Notice that the predicted frame also contains the corresponding probability map of the next waypoint for each robot. Similar to the single-path construction, at each iteration the waypoint for each robot is sequentially determined based on the corresponding predicted probability distribution. To avoid the collision between robots, once a waypoint is occupied, such waypoint cannot be taken by other robots at that timestamp (Line 15-20). The new video frame is then updated according to the new robot states of all robots, and appended into the video clip (Line 27-28). The procedure is repeated until all robots reach their goal regions (Line 29-34).

\section{Results}

\subsection{Dataset}
We demonstrate the effectiveness of STP-Net on three benchmark tasks. The first two tasks are path planning on 2D maps of size $64\times 64$, and the third one is on 3D maps of size $20 \times 20 \times 20$. Specifically, the first type of maps is in the format of random forests, where $15$ obstacles of different shapes are randomly placed in different positions on 2D spaces. The second type is maze-like environments, which are generated using a randomized depth-first search. For the 3D environments, we sample $8$ out of a total of $11$ obstacle blocks with different shapes and place them randomly. For each training task, a feasible, near-optimal path is generated using RRT*. As described in Section~\ref{sec:prob}, each waypoint, the corresponding workspace and the fixed goal state, can be transformed together into a video frame, and hence each training data is interpreted as a video clip. Notice that to facilitate batch training, the video clips of different tasks are all padded into the same length by repeating their last frames, so that all the training data share the same dimensions. 

For both 2D and 3D planning tasks, we train STP-Net over $100$ workspaces, each of which contains $800$ ground truth video clips. The evaluation is performed on both seen and unseen environments. The set of seen environments consists of all the environments used during the training procedure but with $200$ different pairs of start/goal configuration in each environment. The set of unseen environments is composed of $20$ new workspaces which are not used in the training, and each of them contains $400$ different start/goal pairs. We report the accuracy and path cost by averaging over all the computed paths for new tasks from seen/unseen environments.

For the multi-path planning tasks, the data are prepared using a similar procedure. In particular, the video frames of each multi-robot task are generated by stacking the start/goal frame channels of five different tasks under the same environment and at the same timestamp.

\begin{figure}[t]
\vspace{1mm}
  \centering
  \subfigure[$n=17,t=0.15s$]{
    \label{fg:2drf_a}
    \includegraphics[height=1.05in]{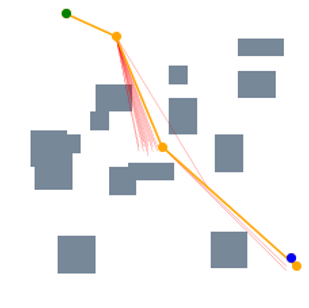}}
  \subfigure[$n=0,t=0.22s$]{
    \includegraphics[height=1.05in]{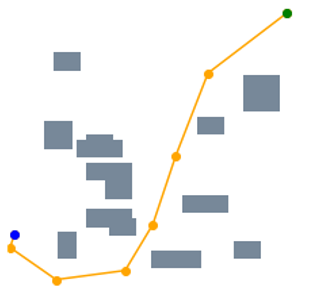}}
  \subfigure[$n=18,t=0.14s$]{
    \includegraphics[height=1.05in]{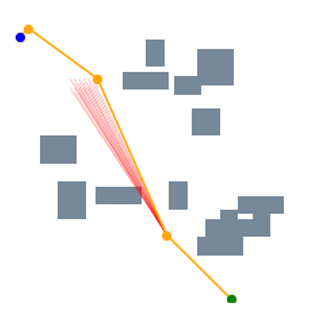}}
  \caption{STP-Net in 2D Random Forest environments. The waypoint, start and goal are marked as orange, green and blue, respectively. Here, $n$ denotes the number of visited waypoints that do not compose the final path solution, and $t$ denotes the time cost of computing the path.}
  \label{fg:2drf}
\end{figure}

\begin{figure}[t]
\vspace{-2mm}  
\centering
  \subfigure[$n=6,t=0.34s$]{
    \label{fg:2dmz_a}
    \includegraphics[height=1.05in]{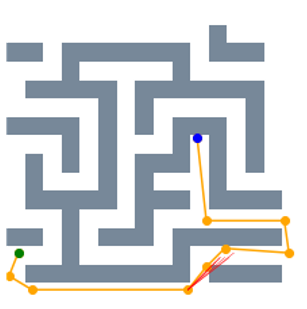}}
  \subfigure[$n=8,t=0.37s$]{
    \label{fg:2dmz_b}
    \includegraphics[height=1.05in]{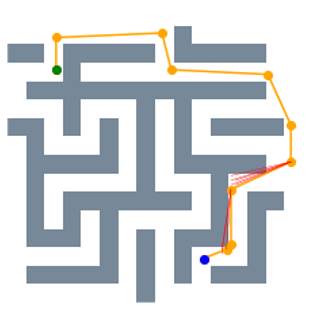}}
  \subfigure[$n=13, t=0.43s$]{
    \label{fg:2dmz_c}
    \includegraphics[height=1.05in]{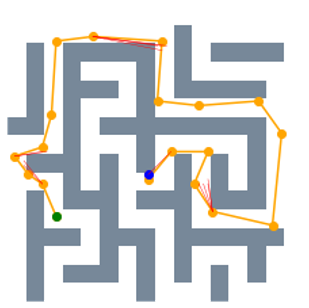}}
  \caption{Visualization of STP-Net path solutions in 2D Maze environments.}
  \label{fg:2dmz}
\end{figure}

\begin{figure}[t]
  \centering
  \subfigure[$n=26, t=0.96s$]{
    \includegraphics[width=0.3\linewidth]{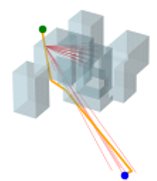}}
  \subfigure[$n=0, t=0.75s$]{
    \includegraphics[width=0.3\linewidth]{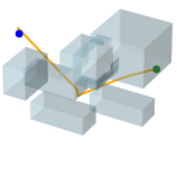}}
  \subfigure[$n=6, t=0.84s$]{
    \includegraphics[width=0.3\linewidth]{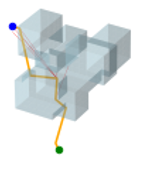}}
  \caption{The paths planned by STP-Net in 3D Random Forest maps.}
  \label{fg:3d}
  \vspace{-3.5mm}
\end{figure}

\vspace{-1mm}
\subsection{Baselines}
We evaluate the performance of STP-Net by comparing it with two classical motion planners (Informed-RRT* and BIT*) and two NN-based ones (MPNet and OracleNet). Considering the original OracleNet has limited performance because of the lack of sufficient map information, for fair comparison we use a performance-enhanced OracleNet-CAE as the baseline, which is equipped with a CAE module (like MPNet does) to better incorporate the encoding vector of the environment into the input.

\begin{table*}[t]
\centering
\vspace{2.5mm}
\begin{tabular}{|l|c|c|c|c|c|c|c|c|c|c|c|c|}
    \hline
    \multirow{2}{*}{Environment}  & \multicolumn{2}{c|}{Informed-RRT*} & \multicolumn{2}{c|}{BIT*} & \multicolumn{2}{c|}{OracleNet-CAE} & \multicolumn{2}{c|}{MPNet} & \multicolumn{2}{c|}{STP-Net (Prob)} & \multicolumn{2}{c|}{STP-Net}\\
    \cline{2-13}
     &  time & cost & time & cost & time & cost & time & cost & time & cost & time & cost \\
    \hline
    \hline
    2D RF Seen & $1.63$ & $82.55$ & $1.14$ & $\textbf{82.23}$ & $1.03$ & $89.55$ &  $0.91$ & $84.79$ &  $0.36$ & $86.30$ & $\textbf{0.17}\pm 0.03$ & $84.17 \pm 4.52$  \\
    \cline{2-13}
    2D RF Unseen & $1.59$ & $83.49$ & $1.10$ & $\textbf{82.80}$ & $1.10$ & $86.34$ &   $0.89$ & $84.50$ & $0.38$ & $85.22$ & $\textbf{0.17} \pm 0.05$ & $83.96 \pm 5.71$ \\
     \hline
    2D Maze Seen & $2.25$ & $121.66$ & $2.06$ & $\textbf{121.34}$ &  $1.30$ & $128.50$&  $0.99$ & $124.48$ & $0.74$ & $126.20$ & $\textbf{0.35} \pm 0.11$ & $122.35 \pm 9.26$ \\
    \cline{2-13}
    2D Maze Unseen & $2.38$ & $120.78$ & $2.17$ & $\textbf{120.15}$ & $1.38$ & $129.32$  & $1.02$ & $124.21$ & $0.77$ & $126.92$ & $\textbf{0.38} \pm 0.10$ & $121.86 \pm 7.94$ \\
     \hline
    3D RF Seen & $4.35$ & $17.91$ & $3.83$ & $\textbf{17.52}$ & $1.71$ & $21.84$&  $1.62$ & $19.68$ & $1.47$ & $20.12$ & $\textbf{0.90} \pm 0.16$ & $18.47 \pm 2.15$ \\
    \cline{2-13}
    3D RF Unseen & $4.20$ & $17.47$ & $3.79$ & $\textbf{17.09}$ & $1.75$ & $21.27$ &  $1.60$ & $19.90$ & $1.55$ & $21.72$ &  $\textbf{0.88} \pm 0.18$ & $18.05 \pm 3.02$ \\
     \hline
\end{tabular}
\caption{Performance comparison between STP-Net and four baseline motion planners on various 2D/3D tasks, w.r.t. the time cost (s) and path cost. Here the path cost is computed according to Eq. \eqref{eq:cost}.}
\label{tb:baseline}
\end{table*}

\vspace{-1mm}
\subsection{Implementation Details}
Our STP-Net model consists of $4$ stacked ConvLSTM layers to learn the hidden representations, and all the hidden dimensions are set to $64$ for 2D tasks and $32$ for 3D tasks. The sizes of convolution kernels in the ConvLSTM are set to $5 \times 5$ for 2D tasks and $5 \times 5 \times 5$ for 3D tasks. At the output end of the model, an additional $1 \times 1$ or $1 \times 1 \times 1$ convolutional layer is allocated to convert the last hidden representation into the output frame that shares the same dimension as the input. In the training phase, ADAM optimizer~\cite{kingma2014adam} is selected with batch size of $128$ for 2D tasks and $64$ for 3D tasks, respectively. The learning rate is set as $0.0003$ and the entire model is trained for $20000$ iterations. The pixels of states are augmented by patches of size $5$ for 2D tasks and $3$ for 3D tasks. Besides, the length of input sequence $J$ from Equation~\ref{eq:train} is set as $5$. The experiments are conducted on a computer equipped with an AMD EPYC 74202P 24-Core Processor and an NVIDIA RTX A6000 GPU. The neural networks are trained and tested using the PyTorch framework. We implement the neural network of STP-Net based on the official code of PredRNN~\cite{wang2017predrnn}. We report the results of Informed-RRT* and BIT* using their C++ OMPL~\cite{sucan2012open} implementation.


\subsection{Ablation Study}
We first perform an ablation study on STP-Net, which justifies the necessity of predicting the static obstacles in future frames. To this end, we modify STP-Net slightly to obtain variant called STP-Net (Prob), which only predicts the probability maps of the next waypoint. From Table~\ref{tb:baseline}, our STP-Net that predicts the 3-channel video frames outperforms STP-Net (Prob) significantly, with respect to both the time cost and path cost. Furthermore, to find a path under 2D Random Forest environments, STP-Net (Prob) visits around four times more waypoints than STP-Net, which causes a greater number of collision detection operations. Given the ablation study, STP-Net (Prob) is omitted from further evaluation.

\begin{figure}[ht]
  \centering
  \subfigure[$t_{multi}=0.27s (t=1.09s)$]{
    \includegraphics[width=0.45\linewidth]{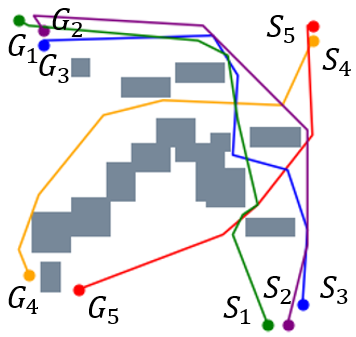}}
  \subfigure[$t_{multi}=0.29s (t=1.16s)$]{
    \includegraphics[width=0.45\linewidth]{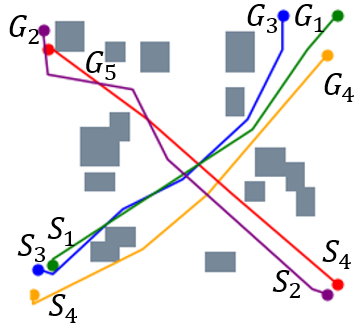}}
  \subfigure[$t_{multi}=1.05s (t=3.27)$]{
    \includegraphics[width=0.50\linewidth]{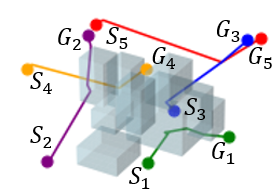}}
  \subfigure[$t_{multi}=1.52s (t=4.83s)$]{
    \includegraphics[width=0.42\linewidth]{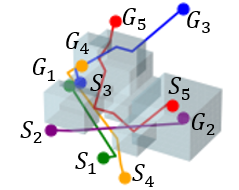}}
  \caption{Our STP-Net on multi-robot path planning tasks under the 2D and 3D Random Forest environments. Here, $t_{multi}$ denotes the time cost of using multi-path construction, and $t$ denotes the total time cost of computing the five paths separately using single-path construction.}
  \label{fg:multi}
\end{figure}

\begin{figure*}[ht]
  \vspace{-6mm}
  \centering
  \subfigure[Path cost.]{
    \includegraphics[height=0.19\linewidth]{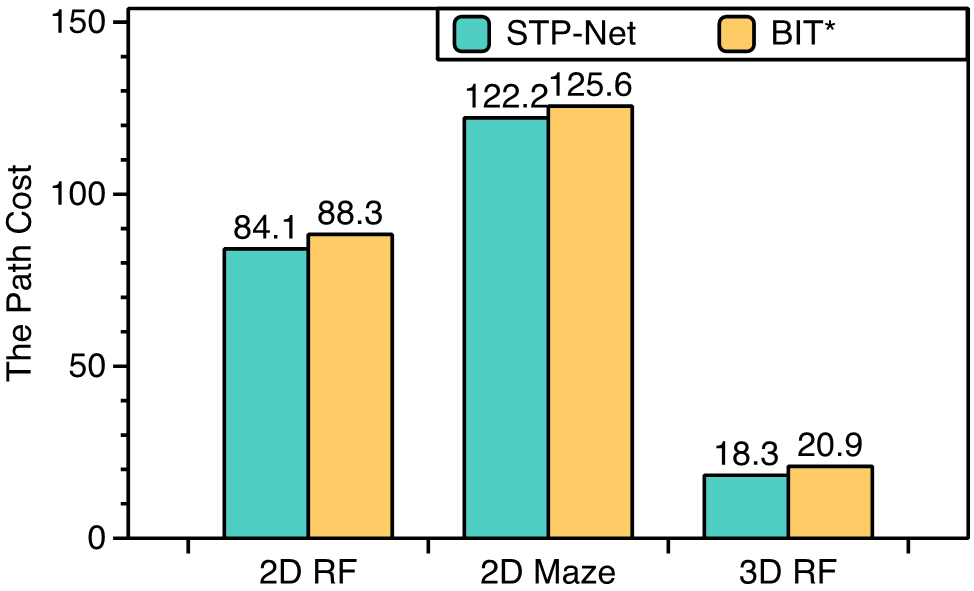}
    \label{fg:pathcost}}
  \subfigure[Number of visited waypoints.]{
    \includegraphics[height=0.19\linewidth]{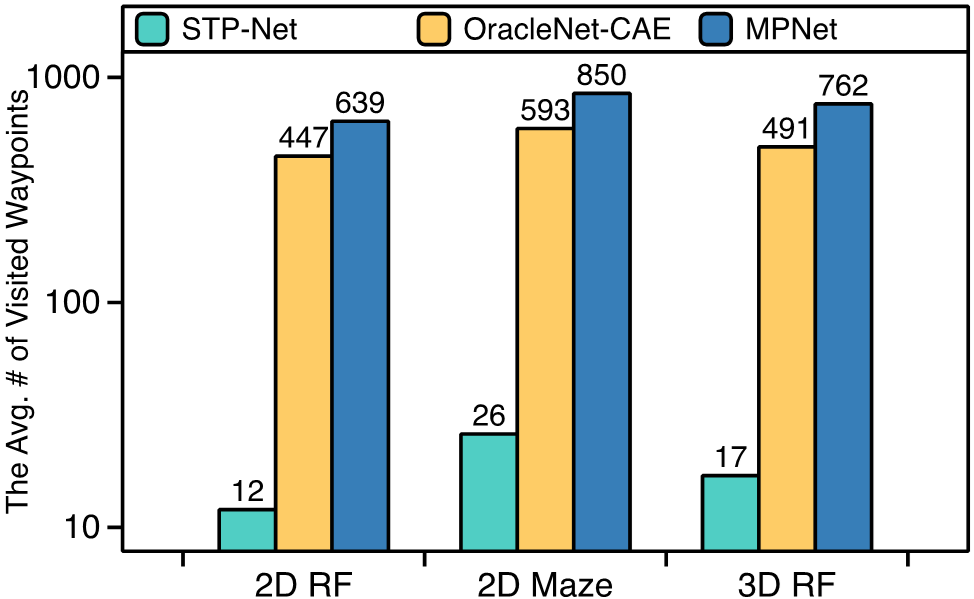}
    \label{fg:samples}}
  \subfigure[Time cost (s).]{
    \includegraphics[height=0.19\linewidth]{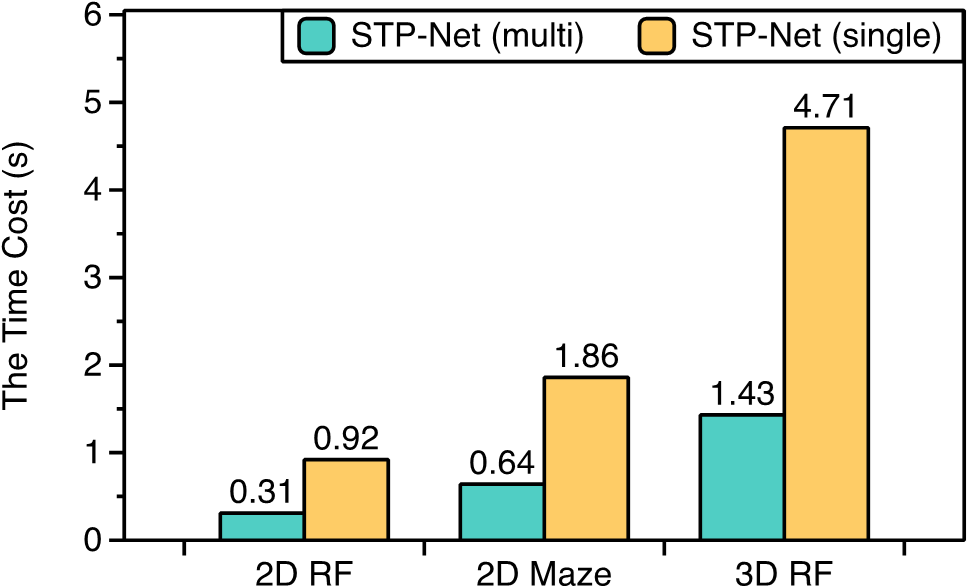}
    \label{fg:multisingle}}
  
    \vspace{-1mm}
  \caption{Left: path costs for STP-Net and BIT* given the same planning time (2D RF: $0.17s$, 2D Maze: $0.35s$, 3D RF: $0.9s$). Middle: the average number of visited waypoints to compute a path. Right: the time cost of computing multi-robot paths in different ways.}
  \label{fg:fig8}
    \vspace{-2mm}
\end{figure*}

\subsection{2D Random Forest Workspaces}
We then evaluate the performance of STP-Net in 2D Random Forest environments. Figure~\ref{fg:process} illustrates how we generate a path step by step via using the predicted frames output from STP-Net model. Initially, the input of the model is a single video frame $\mathcal{X}_0$ (the top-left) containing the full spatial information of the environment plus start and goal states. The model then predicts the next frame $\tilde{\mathcal{X}}_1$ (first row, second column), whose second channel (second row, second column) contains the estimated probabilities of being the next waypoint for each position. Here the brighter yellow pixels indicate higher probabilities. A valid waypoint is determined by selecting the collision-free position with the highest probability. Once the next predicted waypoint is determined, we generate the new observation $\hat{\mathcal{X}}_1$ (third row, second column) and predict the next frame $\tilde{\mathcal{X}}_2$ based on the previous two observations. We repeat this process until the waypoint reaches the goal state at timestamp $t=6$.


From this example, we can further obtain two important observations. \underline{First}, our proposed end-to-end processing strategy, which directly processes the raw environmental map as the input data, can preserve the spatial information very well. As illustrated in the first row (``Frames'') of Figure \ref{fg:process}, each predicted frame, which is essentially the output of STP-Net at the previous timestamp, properly retains all the important spatial information of the map. Such promising properties translate to better planning performance, especially when compared with the current CAE-based encoding strategy that always causes information loss. \underline{Second}, STP-Net learns the collision-free path very well and hence it provides high-quality prediction for the waypoint. Comparing the second row (``Distributions'') and third row (``Decisions'') of Figure \ref{fg:process}, it is seen that in most cases the determined next waypoint lies in the most probable sampling region (the brightest yellow). In other words, this means our neural network model can always predict the collision-free waypoint candidate with high confidence. Such property is very important for improving the planning speed. This is because if the predicted waypoint is further denied after the collision check, the neural planner has to perform more rounds of collision detection until it finds a satisfactory one. Fortunately, the experiments show that the highest probable waypoint candidates predicted by STP-Net are usually collision-free, and thereby naturally saving unnecessary planning cost. Figure~\ref{fg:2drf} further amplifies the importance of high-quality prediction. It is seen that because STP-Net can provide collision-free prediction with very limited failed trials (marked as red lines), the overall planning speed is very fast. 


\subsection{2D Maze \& 3D Random Forest Workspaces}
Figure~\ref{fg:2dmz} visualizes the path search and generation in various maze environments with different task difficulties. It is seen that in such narrow passages-abundant scenarios, STP-Net can still predict the next reachable waypoint in a single shot for most cases. As illustrated in Figure \ref{fg:2dmz}, all the paths can be efficiently found within very limited trials of predictions. The visualization of the generated path of STP-Net in 3D Random Forest workspaces is shown in Figure~\ref{fg:3d}.

\subsection{Comparison with Baselines}
Table~\ref{tb:baseline} compares the performance of our proposed STP-Net with other four baseline motion planners. Experimental results show that all the planners produce $99.8\%-100\%$ accuracy across all the environments. It is seen that STP-Net outperforms the reference approaches with respect to time costs, with the competitive path quality. To be specific, STP-Net is at least $5\times$, $2.6\times$ and $1.8\times$ faster than the other two NN-based methods (MPNet and OracleNet-CAE), on 2D Random Forest, 2D Maze and 3D Random Forest environments, respectively. Simultaneously, STP-Net achieves higher accuracy than MPNet and OracleNet-CAE on all the 2D/3D tasks and regardless whether environments have been seen. 
While it is expected that anytime methods like BIT* should produce 
better path if enough time is granted, as shown in Table~\ref{tb:baseline}, when BIT* is given the same time budget as required by STP-Net, STP-Net produces higher quality paths, which can be observed in Figure~\ref{fg:pathcost}.

Figure~\ref{fg:samples} summarizes the average numbers of predicted waypoints (including the selected ones and the denied ones due to collision) for all the NN-based motion planners. A smaller number here indicates a better prediction produced by a neural network. STP-Net requries at least 22 times fewer predictions of waypoints as compared to the existing learning-based solutions across different planning tasks.



\subsection{Multi-Robot Path Planning}
We further evaluate the performance of STP-Net on five-robot path planning problems. Figure~\ref{fg:multi} shows that STP-Net can produce near-optimal paths for five robots simultaneously with much less time as compared to computing all the paths separately. Figure~\ref{fg:multisingle} compares the time costs of  these two different approaches. It is seen that STP-Net demonstrates strong scalability for multi-robot path planning with three times faster planning speed than the separate single-path construction solution.

\section{Conclusion}
This paper proposes STP-Net, a video prediction-inspired neural motion planner. Empirical evaluations show that, with nearly 100\% accuracy and competitive path quality, STP-Net achieves much faster planning speed than the existing classical and NN-based motion planners. STP-Net also shows good scalability on multi-robot path planning tasks. Thorough evaluation suggests that STP-Net is a very promising NN-based near-optimal motion planner for both single- and multi-robot planning tasks that also runs very fast. 

In future works, we plan to further enhance STP-Net and apply it to challenging motion planning problems in higher dimensions. In particular, we intend to integrate the NN-based STP-Net planning pipeline together with NN-based collision checking to deliver a complete and scalable NN-based motion planning architecture.

\bibliographystyle{unsrt}
\bibliography{reference}

\end{document}